\documentclass{article}

\usepackage[final]{neurips_2024_ml4ps}
\usepackage{graphicx}
\usepackage{mathtools}

\usepackage[utf8]{inputenc} 
\usepackage[T1]{fontenc}    
\usepackage{hyperref}       
\usepackage{url}            
\usepackage{booktabs}       
\usepackage{amsfonts}       
\usepackage{nicefrac}       
\usepackage{microtype}      
\usepackage{xcolor}         

\title{Domain Adaptation of Drag Reduction Policy to Partial Measurements}

\author{%
  Anton Plaksin \\
  Imperial College London\\
  London, UK \\
  \texttt{aplaksin@imperial.ac.uk} \\
  \And
  Georgios Rigas \\
  Imperial College London \\
  London, UK \\
  \texttt{g.rigas@imperial.ac.uk} \\
}

\begin{document}

\maketitle

\begin{abstract}
Feedback control of fluid-based systems poses significant challenges due to their high-dimensional, nonlinear, and multiscale dynamics, which demand real-time, three-dimensional, multi-component measurements for sensing.  While such measurements are feasible in digital simulations, they are often only partially accessible in the real world. In this paper, we propose a method to adapt feedback control policies obtained from full-state measurements to setups with only partial measurements. Our approach is demonstrated in a simulated environment by minimising the aerodynamic drag of a simplified road vehicle. Reinforcement learning algorithms can optimally solve this control task when trained on full-state measurements by placing sensors in the wake. However, in real-world applications, sensors are limited and typically only on the vehicle, providing only partial measurements. To address this, we propose to train a Domain Specific Feature Transfer (DSFT) map reconstructing the full measurements from the history of the partial measurements. By applying this map, we derive optimal policies based solely on partial data. Additionally, our method enables  determination of the optimal history length and offers insights into the architecture of optimal control policies, facilitating their implementation in real-world environments with limited sensor information.
\end{abstract}

\section{Introduction}

Feedback control of fluid-based engineering systems offers significant potential to improve energy efficiency in key sectors such as energy and transport, driving progress toward net-zero emissions. Recent breakthroughs in machine learning have greatly advanced the modeling, optimization, and control of fluid flows \citep{Duriez_2016,Brunton_2020,Rabault_2020,Garnier_2021,Sharma_2023}. Despite these advancements, designing  feedback control remains difficult due to the nonlinear, multiscale nature of fluid dynamics. Optimal control laws typically rely on a dense set of spatio-temporally resolved measurements—a requirement that is feasible in simulations but challenging to meet in real-world applications. In practice, sensor placement is often limited by technical constraints, resulting in only partial measurements being available. In this paper, we propose a framework to adapt feedback control policies, originally developed using full-state measurements, to realistic scenarios where only partial measurements are accessible due to real-world limitations.

We demonstrate our approach in a simulated environment aimed at reducing the aerodynamic drag of road vehicles \citep{Sudin_2014,Choi_2014}. The simulations are conducted in two-dimensional direct numerical simulation (DNS) environments, where the flow dynamics are obtained by numerically solving  the Navier-Stokes equations, and the vehicle is modeled as a simplified 2D bluff body with square geometry. Recent numerical studies \citep{Rabault_2019a,Rabault_2019b,Tang_2020,Paris_2021,Li_2022,Chen_2023} have successfully employed Reinforcement Learning (RL) \citep{Sutton_2018} algorithms to derive feedback control policies in such environments, demonstrating that jet actuators can 
stabilize the unsteadiness in the wake behind the body when full-state measurements in the wake are available. However, in realistic scenarios, control policies rely on the history of measurements on the vehicle (partial measurements), which presents a more complex optimization challenge, as in \cite{Xia_2024}.

In our paper, we address the intermediate scenario where full measurements behind the vehicle are available during the training stage, but only partial measurements are accessible during deployment. To bridge this gap, we propose a supervised learning method of Policy Domain Adaptation (PDA), which leverages a Domain-Specific Feature Transfer (DSFT) map \citep{Wei_2018} to reconstruct full measurements from the history of partial measurements. This approach enables us to achieve the following results:
\begin{itemize}\setlength\itemsep{0.0em}
    \item Taking the composition of the trained DSFT map and the optimal policy for the full measurement case, we obtain the optimal policy for the partial measurement case (see Fig.~\ref{fig:results}a).
    \item Without training a new policy, we analyze the PDA performance depending on the measurement history length and find the minimum length giving the optimal result (see Fig.~\ref{fig:results}b).
    \item Our research provides insights into the architecture of optimal policies for the partial measurement case (see Discussion and Limitations).
\end{itemize}

\begin{figure*}[t]
\centering
\begin{minipage}{.49\textwidth}
\centering
\includegraphics[width=\textwidth]{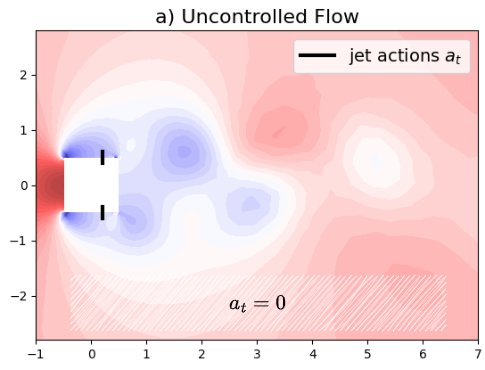}
\end{minipage}
\begin{minipage}{.49\textwidth}
\centering
\includegraphics[width=\textwidth]{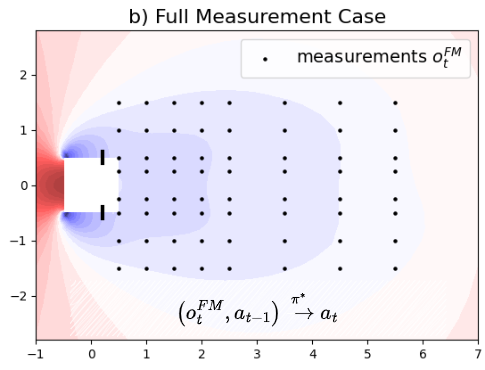}\label{fig:main_b}
\end{minipage}
\\[0.2cm]
\centering
\begin{minipage}{.49\textwidth}
\centering
\includegraphics[width=\textwidth]{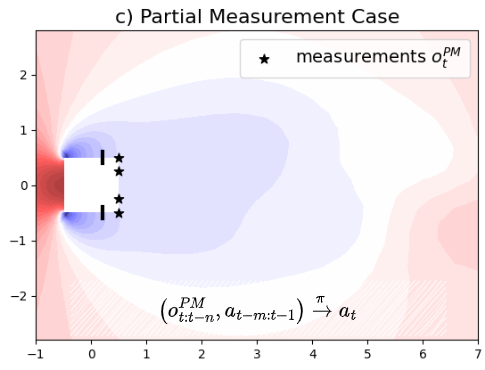}
\end{minipage}
\begin{minipage}{.49\textwidth}
\centering
\includegraphics[width=\textwidth]{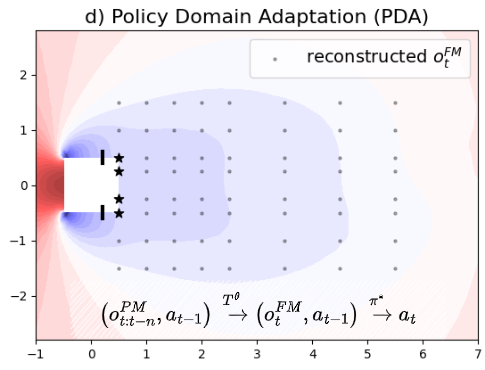}
\end{minipage}
\caption{Flow pressure fields with the jet actuators and measurement locations for the following cases: uncontrolled flow (a); the flow stabilized by the policies trained by TQC algorithm using wake measurements (b) and the measurement history from the body base (c); the flow stabilized by PDA policy (\ref{our_policy}) with $n=48$ and $m=1$ (d). 
}\label{fig:main}
\end{figure*}

\section{Preliminaries and Baselines}

The environment is a 2D direct numerical simulation (DNS) of a laminar flow past a square bluff body with $Re=100$ (see Fig. \ref{fig:main}a). The implementation is taken from \cite{Rabault_2019a,Rabault_2019b} and is based on FEniCS and the Dolfin library \cite{Logg_2012}. The flow dynamics is calculated as a solution $(v_t(\cdot),p_t(\cdot))$ of the Navier-Stokes equation, where $v_t(\cdot)$ is a velocity function and $p_t(\cdot)$ is a pressure function. The flow exhibits vortex shedding in the wake, an instability which increases the drag. The body has two controllable jet actuators on the side edges which influence the vortex shedding. The task is to find a feedback policy that, taking certain measurements as input, reduces the drag of the body by stabilising the vortex shedding.

The discrete-time task can be considered within 
the RL paradigm \citep{Sutton_2018} as a Markov Decision Process (MDP) $(\mathcal{S}, \mathcal{A}, \mathcal{P}, \mathcal{R}, \gamma)$, where $\mathcal{S}$ is the space of all possible states $s_t= (v_t(\cdot),p_t(\cdot),a_{t-1})$, $\mathcal{A}$ is the space of two-dimensional actions $a_t \in [-a_*, a_* ]^2$ defining a smooth change of the jet actuators' force, $\mathcal{P}$ is a transition function giving the next state $s_{t+1} =(v_{t+1}(\cdot),p_{t+1}(\cdot),a_t)$ by the current state $s_t=(v_{t}(\cdot),p_{t}(\cdot),a_{t-1})$ and the current action $a_t$ according to the Navier-Stokes equation, $\mathcal{R}$ is a reward function giving the negative drag coefficient $r_t$, $\gamma$ is a discount factor.
Our aim is to find a policy maximizing the expectation of the total reward $\sum_{t=0}^\infty \gamma^t r_t$.

It is important to note that the state space of this  task is infinite-dimensional, making a direct solution highly challenging. However, previous studies \citep{Chen_2023,Xia_2024} have demonstrated that optimal performance can be achieved within the class of policies $\pi(o^{FM}_t, a_{t-1})$, where $o^{FM}_t$ represents pressure measurements at specific discrete locations in the wake (see Fig. \ref{fig:main}b). In this paper, we replicate these findings and use the optimal policy $\pi^*(o^{FM}_{t},a_{t-1})$ as a baseline target for our approach (represented by the green line in Fig. \ref{fig:results}a).

Although such policies solve the task optimally in simulations, learning them in real-world environments may be technically unfeasible due to the inability to access $o^{FM}_t$. A more realistic task is to find a policy that achieves optimal performance based only on pressure measurements obtained on the body base $o^{PM}_t$ (see Fig. \ref{fig:main}c). Such problem can be considered (see, e.g., \cite{Bertsekas_2012}) as a Partial Observed MDP $(\mathcal{S}, \mathcal{A}, \mathcal{P}, \mathcal{R}, \Omega, \mathcal{O}, \gamma)$, where $\Omega$ is a set of such measurements $o^{PM}_t$ and $\mathcal{O}$ is 
a function given $o^{PM}_t$ by $s_t$. The paper of \cite{Xia_2024} tackled  this task by directly applying RL algorithms to find a policy $\pi(o^{PM}_{t-n:t}, a_{t-m:t-1})$, where $o^{PM}_{t-n:t}$ and $a_{t-m:t-1}$ denote the measurement history $(o^{PM}_{t-n},o^{PM}_{t-n+1},\ldots,o^{PM}_{t})$ and the action history $(a_{t-m},a_{t-m+1},\ldots,a_{t-1})$, respectively. We replicate this approach to establish a baseline policy, which we aim to match or surpass (see red line in Fig. \ref{fig:results}a).

\section{Policy Domain Adaptation}

Similar to \cite{Xia_2024}, our goal is to identify the optimal policy $\pi^*(o^{PM}_{t-n:t}, a_{t-m:t-1})$, but we introduce an alternative approach. Specifically, we assume that the optimal policy  $\pi^*(o^{FM}_t,a_{t-1})$ is already known--this could be obtained, for example, through simulations or using additional measurements that are only available during the training phase but need to be removed at deployment. Additionally, we assume that we have collected a trajectory dataset  $D = \{o^{PM}_{t,i}, o^{FM}_{t,i}, a_{t,i}\}_{t\in\overline{0,T}, i\in\overline{1,k}}$, where $t$ is a step number and $i$ is a trajectory number. Following the DSFT approach  \citep{Wei_2018}, our aim is to establish a map $T$ and select $n$, $m$ such that
$$
T\big(o^{PM}_{t-n:t,i}, a_{t-m:t-1,i}\big) \approx o^{FM}_{t,i},\quad t \in \mathbb{T} = \{\max\{n,m\},\ldots,T\},\quad i \in \mathbb{I} = \{1,\ldots,k\}.
$$
To achieve this, we approximate this map using a neural network $T^\theta$, which is trained by minimising
\begin{equation}\label{loss}
Loss(\theta) = \mathbb{E}_{i \sim \mathbb{I},\ t \sim \mathbb{T}}\ \big\|T^\theta\big(o^{PM}_{t-n:t,i}, a_{t-m:t,i}\big) - o^{FM}_{t,i}\big\|^2 \to \min\limits_{\theta}
\end{equation}
for any m and n without interacting with the environment. The result of our PDA is the policy
\begin{equation}\label{our_policy}
\pi^*\big(o^{PM}_{t-n:t},a_{t-m:t-1}\big) = \pi^*\big(T^\theta\big(o^{PM}_{t-n:t,i}, a_{t-m:t-1,i}\big),a_{t-1}\big). 
\end{equation}

\section{Experiments}

\begin{figure*}[t]
\includegraphics[width=0.48\textwidth]{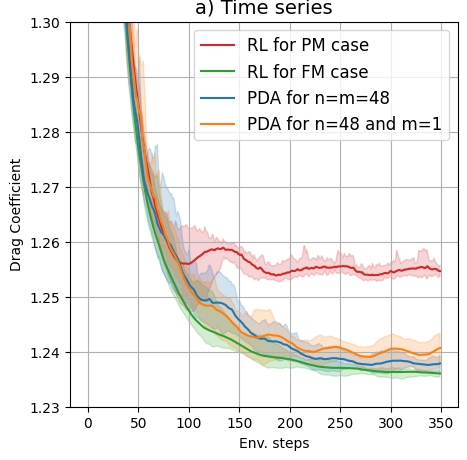}
\includegraphics[width=0.48\textwidth]{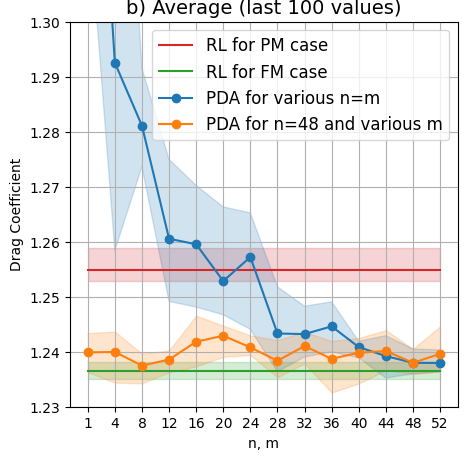}
\caption{Time series and an average of the last 100 values of the Drag Coefficient (negative reward) for the four policies under consideration. Bold lines (left figure) and dots (right figure) indicate the average drag coefficient over 65 runs. Shaded regions illustrate the min-max areas.}\label{fig:results}
\end{figure*}

In this section, we compare the performance of our PDA approach against the baseline cases and analyze its sensitivity to the lengths of the measurement and action histories. Before diving into the performance analysis, it is important to highlight the following key details that contribute to the improved performance of our approach:
\begin{itemize}\setlength\itemsep{0.0em}
    \item Following \cite{Xia_2024}, we apply the TQC algorithm \cite{Kuznetsov_2020} to obtain an optimal policy $\pi^*(o^{FM}_t,a_{t-1})$, but we approximate it with a neural network without hidden layers, i.e. $\pi^{A,B} (o^{FM}_t,a_{t-1}) = a_* \tanh(A o^{FM}_t + B a_{t-1})$, where $A$ and $B$ are parameter matrices. We found this class of policies is sufficient to obtain optimal performance and more robust to approximations, improving the results of our PDA approach (Appendix \ref{appendix:policy}).
    \item We use a neural network with one hidden layer for $T^\theta$. We found a linear map performed worse, and the two hidden layer networks did not improve results (Appendix~\ref{appendix:map}).
\end{itemize}

Fig. \ref{fig:results}a shows the drag coefficient  over time for the trained policies we considered. The corresponding flow fields are shown in Fig. \ref{fig:main}. All policies  converge to a reduced drag state. However, the drag achieved by $\pi(o^{PM}_{t-n:t}, a_{t-m:t-1})$ trained by the RL algorithm directly (red line) is significantly suboptimal compared to the drag of the optimal policy $\pi^*(o^{FM}_t,a_{t-1})$  (green line) and of the policies obtained by our PDA approach (blue and green lines). 

We also study the dependence of the drag on the measurement history length $n$ and the action history length $m$ in Fig. \ref{fig:results}b. Our results indicate that, in general, increasing the measurement history length improves the performance of the PDA-derived policy (blue line), bringing it closer to the optimal value (green line), particularly when $
n=m=48$. Interestingly, the action history does not significantly impact the performance (orange line).

\section{Discussion and Limitations}

From the above results, the following valuable insights can be drawn for future studies:

\begin{itemize}\setlength\itemsep{0.0em}
    \item The studies of \cite{Mao_2022,Xia_2024} consider policies with action history which could indicate that the controlled process is associated with MDP with delay \cite{Katsikopoulos_2003}, also discussed in \cite{Garnier_2021}. However, our optimal policy without the action history (see the orange line in Fig. \ref{fig:results}a) calls into question the necessity to incorporate delay in the formalization of similar drag reduction tasks.
    \item Recently, Partially Observed MDPs have been successfully solved using complex models of data sequence processing (see, e.g., \cite{Chen_2021,Hafner_2024}). Nonetheless, the use of such models for the problems under consideration seems redundant, taking into account the optimal policy we found (\ref{our_policy}), which is a network with only one hidden layer.
    \item We establish that the class of the simplest policies (without hidden layers) is not sufficient to obtain the optimal result, which challenges the effectiveness of linear controllers in partially observable environments. 
\end{itemize}

The primary limitation of our PDA approach arises from the assumption that full measurements are available during the training stage, which is achievable in controlled settings such as digital twin environments or wind-tunnel laboratories. Further research could focus on assessing the robustness of sim-to-real or lab-to-real policy transfer, exploring the performance of policies trained in such environments when deployed in real-world conditions.

\acksection
This research has been funded through the UKRI AI for Net Zero grant "Real-time digital optimisation and decision making for energy and transport systems" (EP/Y005619/1).

\bibliography{paper}
\bibliographystyle{abbrvnat}

\newpage
\appendix
\onecolumn
\section{Appendix}\label{appendix:policy}

In our research, we use TQC algorithm \cite{Kuznetsov_2020} from Stable Baselines3 \cite{stable-baselines3} with hyperparameters mostly taken from \cite{Xia_2024}:

\begin{table}[h]
\centering
\begin{tabular}{ |p{3cm}||p{3cm}| }
\hline
Parameters              & TQC \\[0.1cm]
\hline 
environment steps           & $\approx$ 3.5e6 \\[0.1cm]
parallel environments       & 65 \\[0.1cm]
learning rate               & 1e-4 \\[0.1cm]
batch size                  & 128 \\[0.1cm]
smooth param. $\tau$        & 5e-3 \\[0.1cm]
gradient steps              & 40 \\[0.1cm]
discount factor $\gamma$    & 0.99 \\[0.1cm]
$Q$-model                   & [512,512,512] \\[0.1cm]
$\pi$-model                 & [] or [512,512,512] \\[0.1cm]
\hline
\end{tabular}
\end{table}

We consider two types  of neural networks as a policy model: 
\begin{itemize}
    \item $\pi^*$ is an optimal policy without hidden layers;
    \item $\pi^*_+$ is an optimal policy with three hidden layer with 512 neurons as in \cite{Xia_2024}.
\end{itemize}

\begin{figure*}[h]
\centering
\begin{minipage}{.48\textwidth}
\centering
\includegraphics[width=\textwidth]{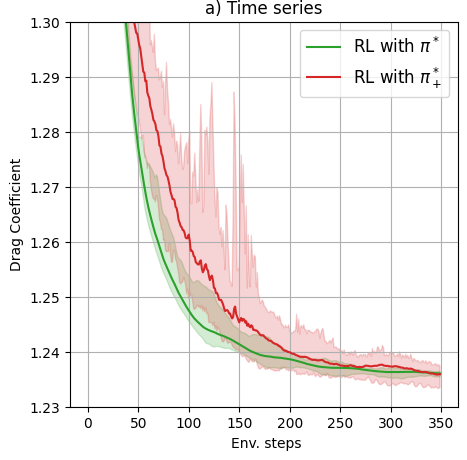}
\end{minipage}
\begin{minipage}{.48\textwidth}
\centering
\includegraphics[width=\textwidth]{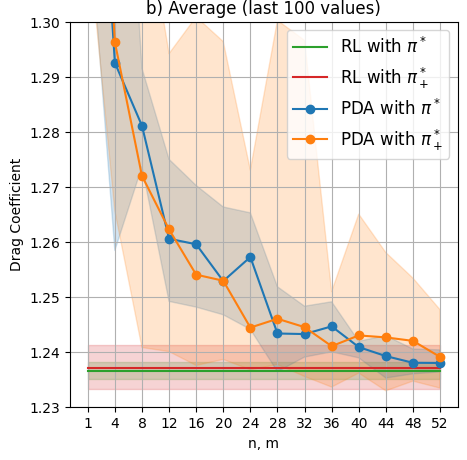}
\end{minipage}
\caption{Time series and average of the last 100 values of the Drag Coefficient (negative reward) for the optimal policy without hidden layers $\pi^*$ and the optimal policy with three hidden layers $\pi^*_+$. Bold lines (left figure) and dots (right figure) indicate the average drag coefficient over 65 runs.}\label{fig:policies}
\end{figure*}

We found both policies are capable of producing optimal performance, but the policy $\pi^*$ appears to be more robust to input imprecision because its results (see Fig. \ref{fig:policies}a) and results of composition~(\ref{our_policy}) with this policy (see Fig. \ref{fig:policies}b) are more stable with respect to runnings.

\newpage
\section{Appendix}\label{appendix:map}

We solve optimization problem (\ref{loss}) with following hyperparameters:

\begin{table}[h]
\centering
\begin{tabular}{ |p{3cm}||p{3cm}| }
\hline
Parameters                  & PDA \\[0.1cm]
\hline 
number of epochs            & 10000 \\[0.1cm]
optimization method         & Adam \\[0.1cm]
learning rate               & 1e-3 \\[0.1cm]
batch size                  & 10000 \\[0.1cm]
dataset normalization       & z-scale \\[0.1cm]
$T$-model                   & [] or [128] or [64,64] \\[0.1cm]
\hline
\end{tabular}
\end{table}

We consider tree type of neural networks to approximate the map $T$: 
\begin{itemize}
    \item the model $T^\theta_-$ without hidden layers (linear);
    \item the model $T^\theta$ with one hidden layer with 128 neurons;
    \item the model $T^\theta_+$with two hidden layer with 64 neurons;
\end{itemize}
\begin{figure*}[h]
\centering
\begin{minipage}{.48\textwidth}
\centering
\includegraphics[width=\textwidth]{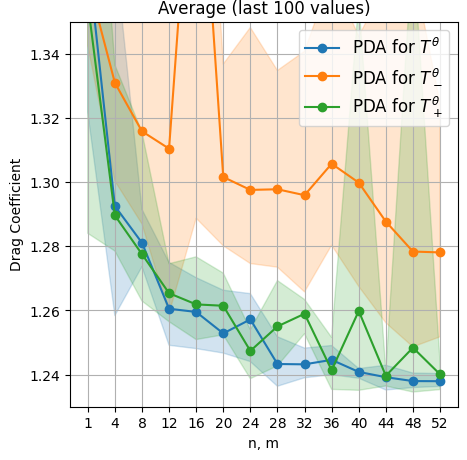}
\end{minipage}
\caption{Average of the last 100 values of  the Drag Coefficient  for the policies obtained by PDA approach as composition~(\ref{our_policy}), where the policy $\pi^*$ is fixed, but the map $T^\theta$ is approximated by the models $T^\theta_-$, $T^\theta$, and $T^\theta_+$. Dots indicate the average drag coefficient over 65 runs.}\label{fig:maps}
\end{figure*}

We found (see Fig. \ref{fig:maps}) the linear map $T^\theta_-$ gets results that are quite far from the optimal, while deeper models $T^\theta$ and $T^\theta_+$ achieve optimal performance and generally show similar results. Nonetheless, note that the model $T^\theta$ is more stable than $T^\theta_+$ over runnings, which may be due to overfitting of the deeper model $T^\theta_+$ on the training data and consequently reducing its generalization ability.

\end{document}